# Unsupervised Domain Adaptation with Similarity Learning

Pedro O. Pinheiro
Element AI
Montreal, QC, Canada
pedro@elementai.com

## Abstract

*The objective of unsupervised domain adaptation is to leverage features from a labeled* source *domain and learn a classifier for an unlabeled* target *domain, with a similar but different data distribution. Most deep learning approaches to domain adaptation consist of two steps: (i) learn features that preserve a low risk on labeled samples (source domain) and (ii) make the features from both domains to be as indistinguishable as possible, so that a classifier trained on the source can also be applied on the target domain. In general, the classifiers in step (i) consist of fully-connected layers applied directly on the indistinguishable features learned in (ii). In this paper, we propose a different way to do the classification, using similarity learning. The proposed method learns a pairwise similarity function in which classification can be performed by computing similarity between prototype representations of each category. The domain-invariant features and the categorical prototype representations are learned jointly and in an end-to-end fashion. At inference time, images from the target domain are compared to the prototypes and the label associated with the one that best matches the image is outputed. The approach is simple, scalable and effective. We show that our model achieves state-of-the-art performance in different unsupervised domain adaptation scenarios.*

## 1. Introduction

Convolutional Neural Networks (ConvNets) [32] based methods achieve excellent results in large-scale supervised learning problems, where a lot of labeled data exists [30, 27]. Moreover, these features are quite general and can be used in a variety of vision problems, such as image captioning [58], object detection [33] and segmentation [26].

However, direct transfer of features from different domains do not work very well in practice, as the data distributions of domains might change. In computer vision, this problem is sometimes referred to as *domain shift* [52]. The most commonly used approach to transfer learned features is to further modify them through a process called "fine-tuning". In this case, the features are adapted by training the network with labeled samples from the new data distribution. In many cases, however, acquiring labeled data can be expensive.

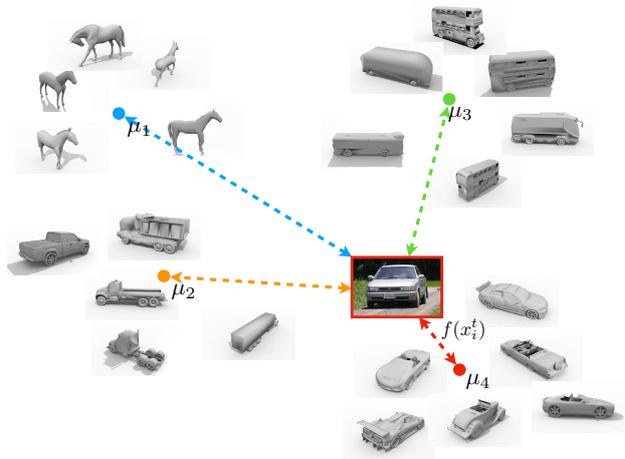

Figure 1: Illustrative example of our similarity-based classifier. A query target-domain image representation is compared to a set of prototypes, one per category, computed from source-domain images. The label of the most similar prototype is assigned to the test image.

*Unsupervised Domain Adaption* deals with the domain shift problem. We are interested in learning representations that are invariant to domains with different data distributions. In this scenario, the machine has access to a labeled dataset (called *source* domain) and an unlabeled one (with a similar but different data distribution, called *target* domain), and the objective is to correctly infer the labels on the latter. Most current approaches are based on deep learning methods and consist of two steps: (i) learn features that preserve a low risk on labeled samples (source domain) and (ii) make the features from both domains to be as indistinguishable as possible, so that a classifier trained on the source can also be applied on the target domain.

Unsupervised domain adaptation can also be used to train models with synthetic data (*e.g.* 3D renders or game engines). Assuming we are able to handle the domain shift problem, we can make use of virtually unlimited number of labeled (synthetic) images and adapt the model to natural images. Indeed, in certain areas of research, such as robotics vision or reinforcement learning, acquiring training sample is very expensive. Training in a synthetic domain and transferring the learned features to real-world environments can be a solution to alleviate the high cost of acquiring and labeling training data.

Theoretical studies in domain adaptation [5, 4] suggest that a good cross-domain representation is one in which the algorithms are not able to identify from which domain the original input comes from. Most current approaches to domain adaptation achieve this goal by mapping cross-domain features into a common space using deep learning methods. This is generally achieved by minimizing some measure of domain variance (such as the Maximum Mean Discrepancy (MMD) [23, 55, 36, 38]), or by matching moments of the two distributions [51].

Another way to deal with the domain variation problem is to make use of adversarial training [1, 15, 53, 54, 7]. In this scenario, the domain adaptation problem is cast as a minimax game between a domain classifier and feature learning (see Figure 2a). A neural network learns features that are, at the same time, as discriminative as possible (in the source domain) and as indistinguishable as possible (among the domains).

In general, the classifier used in the source domain is a simple fully-connected network followed by a softmax over the categories (as in standard supervised learning). While this is, in principle, a good idea (given the representations are trained to be indistinguishable), it leaves the shared representation vulnerable to contamination by noise that is correlated with the underlying shared distribution [47, 7].

In this paper, we propose a different way to do classification, while keeping the adversarial domain-confusion component of [15]. Instead, we propose a *similarity-based classifier* in which each image (from either the source or target domain) is compared to a set of prototypes (or centroides). The label associated to the prototype that best matches the query image is given to it. See Figure 1. Prototypes are vector representations that are representative of each category that appears in the dataset. They are learned at the same time as the image embeddings, and the whole system is backpropagable.

More precisely, we are interested in learning embeddings for the inputs (the source/target domain images), the embedding of each prototype (one per category), and a pairwise similarity function that compares the inputs and the prototypes (see Figure 2b). At the same time, adversarial training is performed on the inputs to force domain-confusion. All these components are learned jointly and in an end-to-end fashion.

At inference time, each embedding prototype is computed a priori (by averaging the embedding over multiple source images). A test image (from target domain) is then compared to each of the prototypes and the label of the nearest prototype is assigned to it.

We show empirically that the proposed similarity-based classification approach is more robust to the domain shift between two datasets. Our method is able to outperform the commonly used classifier and achieves new state of the art in multiple domain adaptation scenarios. We show results in three domain adaptation datasets: *Digits*, which contains digit images from difference domains (MNIST [32], USPS [13] and MNIST-M [15]), *Office-31* [45], which contains images of office objects in three different domains and *VisDA* [42], a large-scale dataset focused on simulation-to-reality shift.

The paper is organized as follows: Section 2 presents related work, Section 3 describes our proposed method and architecture choices, and Section 4 describes our experiments in different datasets. We conclude in Section 5.

## 2. Related Works

**Similarity Learning**: Many previous work focus on learning a similarity measure that is also a metric, like the case of the positive semi definite matrix that defines the Mahalanobis distance. See [3] for a survey.

Multiple authors (*e.g.* [60, 17, 18, 59, 40]) used a learned similarity measure together with nearest neighbors classifier. [48, 46, 10, 44, 9] propose methods in which features are learned along with the metric learning. Chechik *et al.* [9] uses similarity measure in large-scale image search. Similar to us, they define similarity as a bilinear operator but removes the positivity and symmetry constraints to favor scalability. Mensink *et al.* [40] and Snell *et al.* [49], similar to this work, propose to learn metric space in which classification is performed by computing distances to prototype representation of each category (mean of its examples). However, the problem setting and the functions parametrization of these works are very different form ours.

**Unsupervised Domain Adaptation**: Non-deep approaches for unsupervised domain adaptation usually consist of matching the feature distribution between the source and target domain [24, 11]. These methods can roughly be divided into two categories: (i) sample re-weighting [29, 28, 19] and (ii) feature space transformation [41, 22, 2, 50, 31].

Recently, an increasing interest in ConvNets [32] approaches has emerged, due to its ability to learn powerful features. These methods, in general, are trained to minimize a classification loss and maximize domain-confusion. The classification loss is in general computed through a fully-connected network trained on labeled data. The domain-confusion is achieved in two different ways: (i) discrepancy loss, that reduces the shift between the two domains [51, 55, 36, 38, 57, 16] or (ii) adversarial loss, which encourages a common feature space with respect to a discriminator [53, 14, 15, 1, 54].

DeepCORAL [51] achieves the domain-confusion by aligning the second-order statistics of the learned feature representations. In Deep Domain Confusion (DDC) [55], the authors propose a domain-confusion loss based on MMD [23] applied on the final representation of a network. Long *et al*. [36] (DAN) consider a sum of multiple MMDs between several layers. They consider multiple kernels for adapting the representations, achieving more robust results than previous work. In Join Adaptation Networks (JAN) [38], the authors propose an extension to DAN in which they use the joint distribution discrepancy over deep features (instead of their sum). More similar to ours, [57] propose a method that uses hash code to perform classification (and MMD for domain confusion). This method differs from ours in two ways: (i) we consider a different method for domain confusion (adversarial loss instead of MMD) and (ii) we explicitly learn a similarity measure and the prototype embeddings (which brings big improvement in performances as we show in the experiments).

Ganin *et al*. [14] and Ajakan *et al*. [1], and subsequently [15], impose domain confusion through an adversarial objective with respect to a domain discriminator with the reverse gradient algorithm (RevGrad). They treat domain invariance as a binary classification problem, but directly maximizes the loss of the domain classifier by reversing its gradient. This allows the model to learn features that are discriminative in the source domain and indiscriminate with respect to the domain shift. ADDA [54] proposes a general framework for adversarial deep domain adaptation and uses an inverted label GAN [21] loss to achieve the domain confusion. The DSN [7] model divides the space of images into two: one that captures information of the domain, and other that captures information shared between domains.

Certain methods incorporate generative modeling (usually based on GANs [21]) into the feature learning process. Coupled Generative Adversarial Networks [35] consists of a tuple of GANs each corresponding to one of the domains that learns a joint distribution of multi-domain images. In [6], the authors propose a model that generates images (conditioned on source-domain images) that are similar to images from target domain.

Most methods mentioned in this section use the same classifier component: a fully-connected network. In this paper, contrary to them, we use a different classifier based on similarity learning. In our method, we follow the approach of [15] to impose domain-confusion (due to its simplicity and efficiency). We note, however, that our similarity-based classifier could be applied to other approaches mentioned in this section.

## 3. Method

In domain adaptation, we have access to labeled images $\mathbf{X}_s = \{(x_i^s, y_i^s)\}_{i=0}^{N_s}$ drawn from a source domain distribution $p_s(x, y)$ and target images $\mathbf{X}_t = \{(x_i^t, y_i^t)\}_{i=0}^{N_t}$ drawn from a target distribution $p_t(x, y)$. In the unsupervised setting, we do not have any information about the label on the target domain.

We address the problem of unsupervised domain adaptation using a similarity-based classifier. Our model, which we call *SimNet*, is composed of two different components (see Figure 2b): (i) the domain-confusion component, which forces the features of both domains, $f(\mathbf{X}_s)$ and $f(\mathbf{X}_t)$, to be as indistinguishable as possible and (ii) a classifier based on a set of prototypes, $\mu_c$ (one for each category $c \in \{1, 2, ..., C\}$). Both components are trained jointly and in an end-to-end fashion.

This approach is based on the assumption that it exists an embedding for each category such that all the points of the category cluster around it, independent of its domain. Inference is then performed in a test image by simply finding the most semantically similar prototype.

We next describe the proposed similarity-based classifier, followed by the training and inference procedures and implementation details.

### 3.1. Similarity-Based Classifier

Our classifier is composed of $C$ different prototypes, one per category. Each prototype represents a general embedding for a category, incorporating all its variations. We assume that there exists a representation space in which all samples of a given category can be clustered around its corresponding prototype.

Each prototype is represented by a $m$-dimensional vector $\mu_c \in \mathbb{R}^m$ parametrized by a ConvNet $g(\cdot)$, with trainable parameters $\theta_g$. The prototypes are computed by the average representation of all source samples belonging to the category $c$:

$$\mu_c = \frac{1}{|\mathbf{X}^c|} \sum_{x_i^s \in \mathbf{X}^c} g(x_i^s), \qquad (1)$$

where $\mathbf{X}^c$ is the set of all images in the source domain labeled with category $c$. Similarly, the input images (from either domain) are represented as a $n$-dimensional vector $f_i = f(x_i) \in \mathbb{R}^n$, thanks to a ConvNet $f(\cdot)$ parametrized by $\theta_f$.

By leveraging the powerful representations of convolutional neural networks, we propose a simple model that can predict which of the prototypes (and therefore categories) best describes a given input. For this purpose, a similarity metric between images and prototypes is learned, as illustrated in Figure 2b. The similarity between an input image $x_i$ and prototype $\mu_c$ is defined simply as a bilinear operation:

$$h(x_i, \mu_c) = f_i^T \mathbf{S} \mu_c ,  \quad (2)$$

with $\mathbf{S} \in \mathbb{R}^{n \times m}$ being the trainable parameters. $\mathbf{S}$ is an unconstrained bilinear similarity operator, and it is not required to be positive or symmetric.

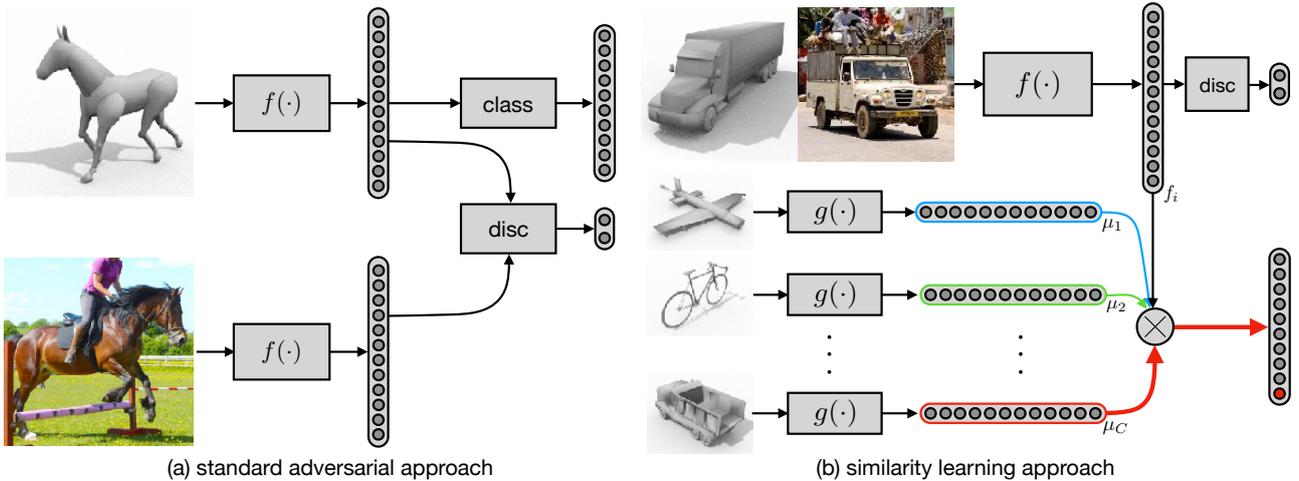

Figure 2: Architectures for adversarial unsupervised domain adaptation. (a) Standard approaches usually consist of two heads: *disc*, which imposes features do be indiscriminate w.r.t. the domain and *class*, that uses a fully-connected network to classify source domain images. (b) Our method keeps the same *disc* component but uses instead a similarity-based classifier. It compares the embedding of an image with a set of prototypes. The label that best matches the image is given. All representations are learned jointly and in an end-to-end fashion.

Note that the two ConvNets, $f$ and $g$, do not share the same parameters. This is particularly important in the domain adaptation scenario, in which the representations (from input and from prototypes) have different roles in the classification. On the one hand, the domain features $f_i$ should, at the same time, be domain invariant and match one of the prototypes $\mu_c$. On the other hand, the embedding prototypes should be as close as possible to the source domain images that represents its category. In the case of single domain classification, it would make sense to use the same network for $f$ and $g$ to reduce the capacity of the model, since there is no shift in the domain. In this case, the model would be similar to Siamese Networks [10].

The model is trained to discriminate the target prototype $\mu_c$ from all other prototypes $\mu_k$ (with $k \neq c$), given a labeled image. We interpret the output of the network as class conditional probabilities by applying a softmax [8] over the bilinear operator:

$$p_\theta(c|x_i, \mu_1, ..., \mu_C) = \frac{e^{h(x_i, \mu_c)}}{\sum_k e^{h(x_i, \mu_k)}} . \quad (3)$$

$\theta = \{\theta_f, \theta_g, \mathbf{S}\}$ represents the set of all trainable parameters of the model. Learning is achieved by minimizing the negative log-likelihood (with respect to $\theta$), over all labeled samples $(x_i, y_i) \in \mathbf{X}^s$:

$$\mathcal{L}_{class}(\theta) = - \sum_{(x_i, y_i)} \left[ h(x_i, \mu_{y_i}) - \log \sum_k e^{h(x_i, \mu_k)} \right] + \gamma \mathcal{R} . \quad (4)$$

$\mathcal{R}$ is a regularization term that encourage the prototypes to encode different aspects of each category. At each training iteration, the prototypes are approximated by choosing a random subset of examples for each class.

The regularizer is modeled as a soft orthogonality constraint. Let $\mathbf{P}_\mu$ be a matrix whose rows are the prototypes, we write the regularization term as:

$$\mathcal{R} = ||\mathbf{P}_\mu^T \mathbf{P}_\mu - \mathbf{I}||_F^2 , \quad (5)$$

where $||\cdot||_F^2$ is the squared Frobenius norm and $\mathbf{I}$ is the identity matrix.

### 3.2. Training and Inference

We are interested in training a classifier on the source domain (labeled samples) and apply it to the target domain (unlabeled samples). To achieve this goal, the model learns features that maximize the domain confusion, while preserving a low risk on the source domain.

The domain-invariant component is responsible to minimize the distance between the empirical source and target feature representation distributions, $f(\mathbf{X}_s)$ and $f(\mathbf{X}_t)$. Assuming this is the case, the classifier trained on the source feature representation can thus be directly applied to the target representation.

Domain confusion is achieved with a domain discriminator $D$, parametrized by $\theta_d$. The discriminator classifies whether a data point is drawn from the source or the target domain, and it is optimized following a standard classification loss:

$$\mathcal{L}_{disc}(\theta, \theta_d) = -\sum_{i=0}^{N_s} \log D(f(x_i^s)) + \\ -\sum_{i=0}^{N_t} \log(1 - D(f(x_i^t))) \ . \quad (6)$$

We achieve domain confusion by applying the Reverse Gradient (RevGrad) algorithm [15], which optimizes the features to maximize the discriminator loss directly. We point the reader to [15] for more details about this component.

The model is trained to jointly maximize the domain confusion (between source and target) and infer the correct category on the source (labeled) samples, through the similarity-based classifier described in the previous section. Therefore, our final goal is to optimize the following minimax objective:

$$\min_{\theta_f, \theta_g, \mathbf{S}} \max_{\theta_d} \mathcal{L}_{class}(\theta_f, \theta_g, \mathbf{S}) - \lambda \mathcal{L}_{disc}(\theta_f, \theta_d) \ , \quad (7)$$

where $\lambda$ is a balance parameter between the two losses. The objective is optimized using stochastic gradient descent with momentum.

At inference time, the prototypes are computed a priori, following Equation 1, and stored in memory. The similarity between a target-domain test image and each prototype is computed, and the label that best matches the query is outputed.

### 3.3. Implementation Details

In our large-scale experiments (*i.e. Office-31* and *VisDA* datasets), the parameters of networks $f$ and $g$ are initialized with a ResNet-50 [27] that was pre-trained to perform classification on the ImageNet dataset [12], and the classification layer is removed. For the *Digits* experiments, both $f$ and $g$ have three convolution layers with $5 \times 5$ kernels and stride 1 (with batch normalization and ReLU non-linearities) with 64, 64 and 128 hidden units, respectively. We included a max-pooling after the first two convolutions.

The discriminator network and the bilinear classifier are initialized randomly from a uniform distribution. We set the balance parameter $\lambda = 0.5$ and the regularization coefficient $\gamma = 0.01$ (we observed that the model is robust to this hyperparameter). We use a learning rate of $10^{-5}$ ($10^{-3}$ for digits), with a weight decay of $10^{-5}$ and momentum of 0.99. Since the similarity matrix and the discriminator are trained from scratch, we set their learning rate to be 10 times that of the other layers.

During training, the images (from both domains) are resized such that the shorter dimension is of size 300 pixels (32 for digits) and a patch of $224 \times 224$ ($28 \times 28$ for digits) is randomly sampled. Each mini-batch has 32 images from each domain. At each training iteration, the prototypes are approximated by picking one random sample for each class. We noticed that the training converges with this design choice and it is more efficient.

The discriminator is a simple fully-connected network. It contains two layers, each of dimension 1024 and ReLU non-linearity, followed by the domain classifier. It receives as input the output of network $f$ and outputs the probability of which domain the input comes from (this probability is, again, modeled by a softmax).

We parametrize the bilinear operation with a low-rank approximation, $\mathbf{S} = \mathbf{U}^T \mathbf{V}$ ($\mathbf{U}, \mathbf{V} \in \mathbb{R}^{n \times m}$, $m = 512$) and Equation 2 becomes:

$$h(x_i, \mu_c) = (\mathbf{U} f_i)^T \cdot (\mathbf{V} \mu_c) \ . \quad (8)$$

This parametrization brings multiple benefits. First, it allows us to control the capacity of the model. Second, it can trivially be implemented in any modern deep learning framework and benefits from an efficient implementation. Finally, it also provides fast inference, as the right side of $h$ is independent of the input image; it can be computed only once and stored in memory.

At inference time for the large-scale experiments, the shorter dimensions of the test image is resized to 300 pixels, as in the training stage. The model is applied densely at every location, resulting in an output with spatial dimensions bigger than 1. The output is averaged over the spatial dimensions to get one-dimensional vector. Because the similarity measure is a bilinear operation, the averaging can be seen as an ensemble over different spatial locations of the test image. In the Digits experiments, we directly forward the ($28 \times 28$ input).

## 4. Experimental Results

We evaluate the performance of our model, SimNet, in three important unsupervised domain adaptation datasets and across different domain shifts. Figure 3 illustrates image samples from different datasets and domains.

In the experiments, we use all labeled source images and all unlabeled target images, following the standard evaluation protocol for unsupervised domain adaptation [15, 38]. All hyperparameters were chosen via transfer cross-validation [61]. We evaluate the performance of all methods with classification accuracy metric.

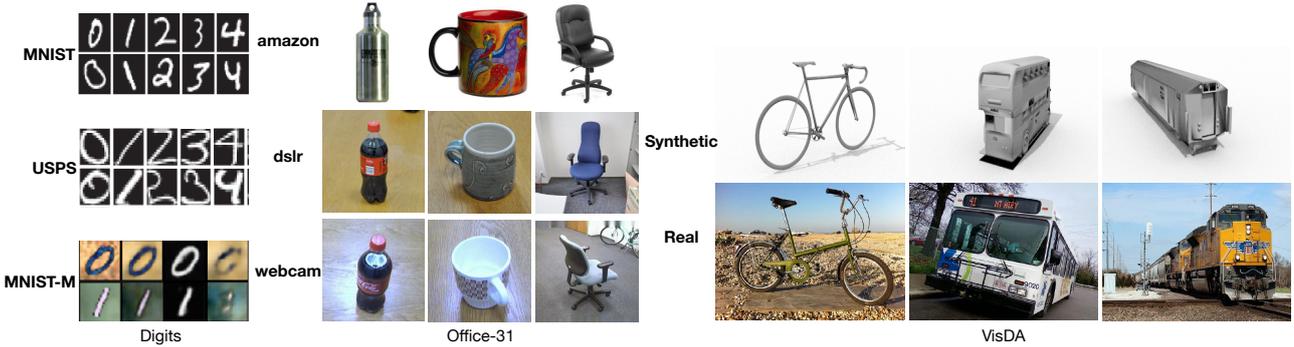

Figure 3: We evaluate our model on unsupervised domain adaptation across different domains in three distinct settings. The first setting is on different digits datasets (MNIST, USPS and MNIST-M). The second setting is over all 6 combinations of domains in Office-31 dataset. The third is on the synthetic-to-real domain shift on VisDA dataset.

|  | USPS to MNIST | MNIST to USPS | MNIST to MNIST-M |
|---|---|---|---|
| RevGrad-ours [15] | 89.9 | 89.1 | 84.4 |
| CoGAN [35] | 89.1 | 91.2 | 62.0 |
| ADDA [54] | 90.1 | 89.4 | – |
| DSN [7] | – | 91.3 | 83.2 |
| AssocDA [25] | – | – | 89.5 |
| PixelDA [6] | – | 95.9 | 98.2 |
| SimNet | 95.6 | 96.4 | 90.5 |

Table 1: Per-class average classification accuracy (%) on the three digits domain shift evaluated.

### 4.1. Digits Results

We evaluate our model in three different digits domain shifts: (i) USPS to MNIST, (ii) MNIST to USPS and (iii) MNIST to MNIST-M, a variation of MNIST for domain adaptation proposed by [15]. In MNIST-M, the samples are created by using each MNIST digit as a binary mask and inverting it with the colors of a background image (made of random crops from BSDS500 dataset [39]). In our experiments, we follow a experimental protocol similar to [7, 6]. For the third experiment, we use the same data augmentation as [25]: we randomly invert MNIST images since they are always white on black, unlike MNIST-M.

Table 1 compares our method with other approaches. We are able to achieve competitive results in all different domain shifts. We note that PixelDA [6] performs extremely well on the third task because it learns a transformation (on pixel level) from one domain to another – and this is precisely how the target domain is generated in this scenario. This, however, is not the case in more complex dataset shift.

### 4.2. Office-31 Results

Office-31 [45] is a standard dataset used for domain adaptation. The dataset contains 4652 images of 31 different categories. The categories are everyday office objects, such as backpack, pen and phone. The images on this dataset belong to three distinct domains: (i) *Amazon* (A), which contains images from the Amazon website, (ii) *DSLR* (D) and (iii) *Webcam* (W), which contains photos taken by a high quality camera and a webcam, respectively (see Figure 3, left). We evaluate the performance of our method in all 6 possible transfer tasks, using the same experimental protocol as [38].

Table 2 compares the performance of our method with other approaches. The first row shows results without any adaptation, where the network is fine-tuned on the source domain and directly applied to the target domain. This serves as a lower bound. Other results are reported from Long *et al*. [38]. For a fair comparison, all models use the same base architecture, ResNet-50. SimNet beats the previous state of the art in most settings, while being similar in the remaining ones.

### 4.3. VisDA Results

VisDA dataset [42] is focused on simulation-to-real domain shift.[1] It contains two very distinct domains: (i) *Synthetic*, which contains synthetic renderings of 3D models from different angles and with different lightning conditions and (ii) *Real*, which contains natural images cropped either from COCO [34] (validation) or Youtube-BB [43] (test) datasets (see Figure 3, right). It contains over 280K images across 12 categories in the combined training, validation and test domains.

**Comparison to other methods:** Table 3 shows the performance of different methods on the VisDA dataset, for the synthetic to real (validation set) domain shift.[2] Once again, all the experiments are computed using ResNet-50 backbone for fair comparison with the models. The results reported are the average per-class accuracy. Results from other models are taken from Long *et al.* [38][3].

---
[1]This dataset also contains segmentation labels, but on this paper we focus on classification task only.
[2]Some non-published results can be found on-line. However, no detail of the methods nor how many tricks were used are mentioned.

|  | A→W | D→W | W→D | A→D | D→A | W→A | average |
|---|---|---|---|---|---|---|---|
| Source Only | 68.4 ± 0.2 | 96.7 ± 0.1 | 99.3 ± 0.1 | 68.9 ± 0.2 | 62.5 ± 0.3 | 60.7 ± 0.3 | 76.1 |
| TCA [41] | 72.7 ± 0.0 | 96.7 ± 0.0 | 99.6 ± 0.0 | 74.1 ± 0.0 | 61.7 ± 0.0 | 60.9 ± 0.0 | 77.6 |
| GFK [20] | 72.8 ± 0.0 | 95.0 ± 0.0 | 98.2 ± 0.0 | 74.5 ± 0.0 | 63.4 ± 0.0 | 61.0 ± 0.0 | 77.5 |
| DDC [55] | 75.6 ± 0.2 | 96.0 ± 0.2 | 98.2 ± 0.1 | 76.5 ± 0.3 | 62.2 ± 0.4 | 61.5 ± 0.5 | 78.3 |
| DAN [36] | 80.5 ± 0.4 | 97.1 ± 0.2 | 99.6 ± 0.1 | 78.6 ± 0.2 | 63.6 ± 0.3 | 62.8 ± 0.2 | 80.4 |
| RTN [37] | 84.5 ± 0.2 | 96.8 ± 0.1 | 99.4 ± 0.1 | 77.5 ± 0.3 | 66.2 ± 0.2 | 64.8 ± 0.3 | 81.6 |
| RevGrad [15] | 82.0 ± 0.4 | 96.9 ± 0.2 | 99.1 ± 0.1 | 79.7 ± 0.4 | 68.2 ± 0.4 | 67.4 ± 0.5 | 82.2 |
| JAN [38] | 85.4 ± 0.3 | 97.4 ± 0.2 | 99.8 ± 0.2 | 84.7 ± 0.3 | 68.6 ± 0.3 | 70.0 ± 0.4 | 84.3 |
| JAN-A [38] | 86.0 ± 0.4 | 96.7 ± 0.3 | 99.7 ± 0.1 | 85.1 ± 0.4 | 69.2 ± 0.4 | 70.7 ± 0.5 | 84.6 |
| SimNet | 88.6 ± 0.5 | 98.2 ± 0.2 | 99.7 ± 0.2 | 85.3 ± 0.3 | 73.4 ± 0.8 | 71.8 ± 0.6 | 86.2 |

Table 2: Classification accuracy evaluation (%) on Office-31 dataset for unsupervised domain adaptation with different domain pairs. All models utilize ResNet-50 as base architecture.

|  | avg. acc. |
|---|---|
| Source Only | 49.51 |
| DAN [36] | 53.02 |
| RTN [37] | 53.56 |
| RevGrad [15] | 55.03 |
| JAN [38] | 61.06 |
| JAN-A [38] | 61.62 |
| RevGrad-ours | 58.62 |
| SimNet | 69.58 |

Table 3: Per-class average classification accuracy (%) on VisDA-val dataset for unsupervised domain adaptation. All the models utilize ResNet-50 as base architecture.

*RevGrad-ours* reports our implementation of the RevGrad method [15] (with a ResNet-50 architecture), in which SimNet shares the adversarial learning component. As RevGrad-ours and SimNet were trained with same training setup and discriminator function, we can disentangle the performance improvements achieved by the proposed similarity-based classifier. SimNet achieves over 10% boost in performance. This result gives us a hint that similarity-based classifiers are more robust than typical fully-connected networks when features are subject to domain invariance.

VisDA is more challenging than Office-31. The objects in the target domain contains more variability and the domain shift between the two domains is much bigger. Results in Table 3 show that the model scales nicely with the data. SimNet achieves a bigger improvement in performance, compared to previously published state of the art [38], when the scale of the dataset increases. These results are achieved without any inference trick (such as model ensembling, multi-scale inference, etc.).

SimNet, when trained with VisDA-test as the target domain and without bells and whistles, achieves a similar performance: 68.66% average per-class accuracy on the test set.

**Model variants:** Table 4 shows the performance of different versions of SimNet in each of the 12 categories of the VisDA-val. As before, *Source Only* shows the results when trained on the source images only (this is a lower bound). In this experiment, we use the proposed similarity-based classifier and we notice that even in the 'source only' scenario it performs better than the fully-connected one: a performance of 49.5% against 46.0%. *Train on Target* shows results when the model is trained with the labels of the target domain (this can be interpreted as an upper bound).

We also show that sharing the weight between the two networks, $f$ and $g$, hurts the performance (*SimNet-f=g*). The features learned by $f$ need to be domain agnostic, while this requirement is not necessary for $g$. Imposing this constraint on $g$ would make the learning more difficult and thus hurt the performance. The results of *SimNet-no-reg* emphasis the importance of the regularizer described in Equation 5. The penalty term imposes the prototypes to be orthogonal, which eases the similarity computation. In *SimNet-152*, we show the results using a ResNet-152 (pre-trained on ImageNet) as base architecture, which pushes the performance even further. This results shows that proposed bilinear classifier is also able to scale with the quality of the features. For more insight on the performance of different categories, we plot different confusion matrices in Figure 4.

**Feature visualization:** Finally, we use t-SNE [56] to visualize feature representations from different domains and at different adaptation stages. Figure 5(a-b) show features from source (blue) and target (red) domains before and after adaptation, respectively. The features become much more domain invariant after adaptation. We observe a strong correspondence in terms of classification performance (on the target domain) and the overlap between the domain distributions.

---
[3]These values were presented by the authors on the conference where the published paper was presented.

| | aero. | bike | bus | car | horse | knife | moto. | person | plant | sktb. | train | truck | avg. |
|---|---|---|---|---|---|---|---|---|---|---|---|---|---|
| Source Only | 67.7 | 36.6 | 48.4 | 68.2 | 76.9 | 5.3 | 65.8 | 38.0 | 72.5 | 29.1 | 82.1 | 3.73 | 49.5 |
| RevGrad-ours | 75.9 | 70.5 | 65.3 | 17.3 | 72.8 | 38.6 | 58.0 | 77.2 | 72.5 | 40.4 | 70.4 | 44.7 | 58.6 |
| SimNet-f=g | 85.7 | 51.7 | 58.5 | 53.4 | 74.5 | 23.9 | 69.7 | 68.9 | 54.5 | 50.9 | 76.8 | 21.2 | 57.5 |
| SimNet-no-reg | 92.1 | 81.3 | 68.7 | 39.8 | 86.4 | 10.2 | 68.4 | 79.8 | 87.5 | 69.8 | 73.1 | 35.9 | 66.1 |
| SimNet | 94.5 | 80.2 | 69.5 | 43.5 | 89.5 | 16.6 | 76.0 | 81.1 | 86.4 | 76.4 | 79.6 | 41.9 | 69.6 |
| SimNet-152 | 94.3 | 82.3 | 73.5 | 47.2 | 87.9 | 49.2 | 75.1 | 79.7 | 85.3 | 68.5 | 81.1 | 50.3 | 72.9 |
| Train on target | 99.5 | 91.9 | 97.3 | 96.8 | 98.3 | 98.5 | 94.1 | 96.2 | 99.0 | 98.2 | 97.9 | 82.3 | 95.8 |

Table 4: Per-class classification accuracy (%) for different variants of SimNet on VisDA-val.

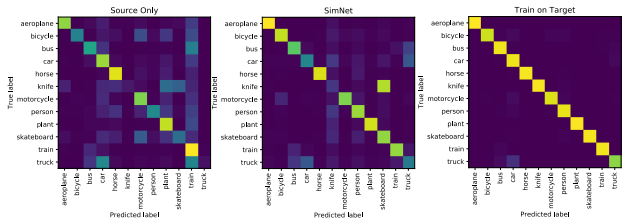

Figure 4: Confusion matrix for source only, SimNet and oracle (trained on target domain) models on the synthetic-to-real adaptation on VisDA-val dataset. Best viewed in color.

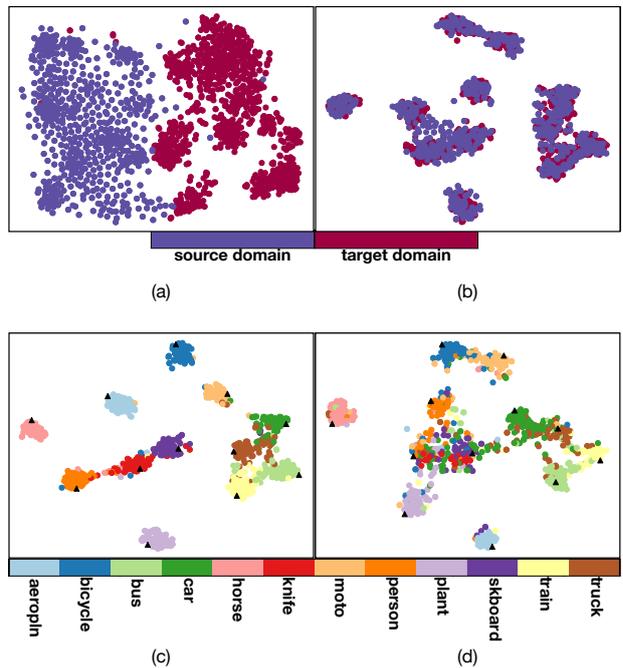

Figure 5: (a-b) Image features from source (blue) and target (red) domains, respectively. The adaptation makes the the distribution of features much closer. (c-d) Image features after adaptation from source and target domain, respectively. Each category is encoded by a color and the prototypes are shown as black triangles. Best viewed in color.

Figure 5(c-d) show features after adaptation from source and target domains, respectively. Each category is encoded by a color and the prototypes are shown as black triangles. As expected, the images on the source domain are separated better than the images from the target domain. We can observe few interesting facts about these plots. First, the similarity measure is well learned for certain categories ('aeroplane','horse', 'plant'). Categories that are similar in terms of appearance and semantics are closer in the t-SNE space: 'bicycle' and 'motorcycle' are very close and 'truck' is located between 'car' and 'bus'. Interestingly, SimNet seems to confuse the categories 'knife' and 'skateboard'.

## 5. Conclusion

In this paper we propose SimNet, a neural network model fur unsupervised domain adaptation using a similarity-based classifier. We show that similarity learning, together with feature learning, can outperform by a large margin the 'standard' fully-connected classifier in the problem of domain adaptation. The method is simple and highly effective. We demonstrate its efficacy by applying it in different unsupervised adaptation problems, achieving new state-of-the-art performance in multiple scenarios. Future works include applying the model in different modalities of domain adaptation (*e.g.* semantic segmentation, object detection) and with different domain discrepancy reduction algorithms (such as MMD and its variants). It would also be interest to explore how the similarity-based classifier can scale to a larger number of categories and training samples.

**Acknowledgements.** I thank Negar Rostamzadeh, Thomas Bosquet and Ishmael Belghazi for helpful discussions and encouragement and Philippe Mathieu and Jean Raby for help with computational infrastructure.